\documentclass[conference]{IEEEtran}
\IEEEoverridecommandlockouts
\usepackage{cite}
\usepackage{amsmath,amssymb,amsfonts}
\usepackage{algorithmic}
\usepackage{graphicx}
\usepackage{textcomp}
\usepackage{xcolor}
\usepackage{multirow}
\def\BibTeX{{\rm B\kern-.05em{\sc i\kern-.025em b}\kern-.08em
    T\kern-.1667em\lower.7ex\hbox{E}\kern-.125emX}}

\usepackage{color, colortbl}
\definecolor{Gray}{gray}{0.8}

\usepackage{comment}
\usepackage{soul}

\usepackage{fancyhdr}
\begin{document}

\title{Play with Emotion:\\Affect-Driven Reinforcement Learning
\thanks{This project has received funding from the European Union’s Horizon 2020 programme under grant agreement No 951911.}
}

\author{\IEEEauthorblockN{
Matthew Barthet,
Ahmed Khalifa,
Antonios Liapis,
Georgios N. Yannakakis\\
\IEEEauthorblockA{Institute of Digital Games, University of Malta, Msida, Malta.\\
Email: \{matthew.barthet, ahmed.khalifa, antonios.liapis, georgios.yannakakis\}@um.edu.mt}}}

\maketitle
\thispagestyle{fancy}

\begin{abstract}
This paper introduces a paradigm shift by viewing the task of affect modeling as a reinforcement learning (RL) process. According to the proposed paradigm, RL agents learn a policy (i.e. \emph{affective interaction}) by attempting to maximize a set of rewards (i.e. behavioral and affective patterns) via their experience with their environment (i.e. \emph{context}). Our hypothesis is that RL is an effective paradigm for interweaving affect elicitation and manifestation with behavioral and affective demonstrations. Importantly, our second hypothesis---building on Damasio's \emph{somatic marker hypothesis}---is that emotion can be the facilitator of decision-making. We test our hypotheses in a racing game by training Go-Blend agents to model human demonstrations of arousal and behavior; Go-Blend is a modified version of the Go-Explore algorithm which has recently showcased supreme performance in hard exploration tasks. We first vary the arousal-based reward function and observe agents that can effectively display a palette of affect and behavioral patterns according to the specified reward. Then we use arousal-based state selection mechanisms in order to bias the strategies that Go-Blend explores. Our findings suggest that Go-Blend not only is an efficient affect modeling paradigm but, more importantly, affect-driven RL improves exploration and yields higher performing agents, validating Damasio's hypothesis in the domain of games.
\end{abstract}


\section{Introduction} \label{sec:introduction}

Affect modeling is predominately viewed from a supervised learning perspective, where a model is trained to link manifestations of affect to ground truth labels provided by human annotators \cite{calvo2010affect}. If the task of affect modeling is instead viewed from a reinforcement learning (RL) lens, the context, the behavioral patterns and the corresponding affective manifestations can be modeled under a common representation. Generated RL agents would then be able to experience the environment (i.e. context), and learn to take actions (i.e. affective interactions) that will maximize a set of rewards (i.e. behavioral and emotional patterns) for the agent. 

While the use of RL in affective computing remains in its infancy, recent studies suggest that the RL paradigm is well suited for affective computing (AC). Specifically, RL is able to retrieve the relationship between affect elicitation, affect manifestation, and affect annotation through common affect-behavioral representations \cite{barthet2021go}. In this paper we investigate the impact of affect in the form of dissimilar reward functions for the generation of agents that model the behavior and affective responses of humans as provided through demonstrations. Importantly, we are inspired by Damasio's \emph{somatic marker hypothesis}---suggesting the critical role of emotion as a \emph{facilitator} of decision-making \cite{damasio1996somatic,bechara2004role}---and we test, for the first time, the effectiveness of affect as a state selection mechanism within the RL paradigm.

\begin{figure}[!tb]
\centering
\includegraphics[width=\columnwidth]{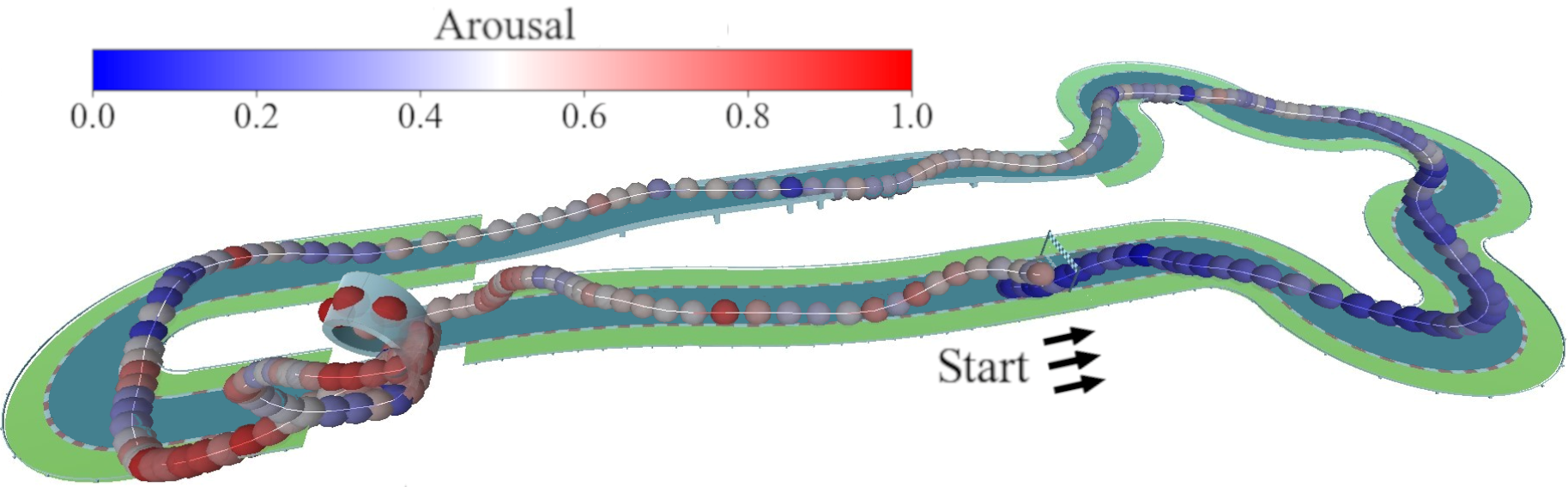}
\caption{Go-Blend: Agent trained to drive on a racing track merely by modeling annotated arousal of expert players.}
\label{fig:teaser}
\end{figure}

Motivated by the lack of comprehensive studies on the effect of emotion for training affective RL agents and guiding decision-making, we build on and extend significantly the Go-Blend framework \cite{barthet2021go}---an implementation of the recent cutting-edge RL algorithm \emph{Go-Explore} \cite{Go-Explore}---that is able to fuse and model behavioral and affective patterns of humans through generative agents. We start our investigations by exploring dissimilar reward functions for modeling behavior and affect. We then test the impact of affect as a state selection mechanism for Go-Blend, thus directly employing affect as the underlying driver of decision-making \cite{damasio1996somatic,bechara2004role}. We test the various implementations of Go-Blend in a fast-paced continuous-controlled real-time rally-driving game; see Fig.~\ref{fig:teaser}. The game features a dataset of over 100 human behavioral (playing) demonstrations with corresponding self-annotated arousal traces.

Our obtained results suggest that all designed reward functions yield affect models that achieve their goal in capturing aspects of arousal accurately. The dissimilar reward functions that cater for aspects of arousal uncertainty and prediction confidence appear to yield dissimilar yet efficient agent behaviors that model annotated arousal with high accuracy. Regarding our experiments employing arousal as a selection mechanism, it seems that affect-based selection is beneficial for boosting the behavior of agents as RL search prioritizes areas with high affect modeling capacity. Such areas are, in turn, helping the algorithm to explore better within the behavioral space of agents.

\section{Background} \label{sec:background}

This section provides a general overview of the use of RL in games and the Go-Explore algorithm (section \ref{sec:background-go-explore}), as well as an overview of affective computing and the current uses of affect for training RL agents (section \ref{sec:background-rl-affect}).

\subsection{RL for Playing Games} \label{sec:background-go-explore}

Reinforcement Learning (RL) is a well established family of machine learning algorithms. When applied to games, RL is usually applied to train agents to play games optimally (i.e. playing to win) and as efficiently as possible \cite{yannakakis2018artificial}. Notable achievements of recent deep RL algorithms for optimal play include reaching super human performance levels in board games such as \emph{Go} \cite{RL-GO} and digital games such as \emph{DotA 2} \cite{RL-Dota}, \emph{StarCraft} \cite{RL-Starcraft}, and \emph{Atari} Games \cite{RL-Atari}. More recent RL frameworks such as \textit{Go-Explore} \cite{Go-Explore} and \textit{BeBold} \cite{Bebold} were developed specifically to tackle hard exploration problems due to \textit{sparse} and \textit{deceptive} reward signals \cite{anderson2018deceptive}.

Given its proven track record for difficult learning tasks such as hard-exploration games \cite{Go-Explore}, text-based games \cite{ammanabrolu2020avoid}, and maze navigation \cite{matheron2020pbcs}, we extend Go-Explore in order to tackle RL for affective computing tasks. In this paper, we focus on Go-Explore's exploration phase, which stores an archive of promising game states along with the trajectories required to return to them in a deterministic setting. In Go-Explore, a trajectory refers to the sequence of actions an agent must follow to reproduce a given state. The original Go-Explore implementation performs exploration by randomly selecting a state from the archive, returning to its game state, conducting a number of random exploratory actions, and storing the discovered states back in the archive based on their associated reward (see section \ref{sec:algorithm}). The result of this exploration phase is a number of promising trajectories which can be used to train agents to behave in stochastic settings.

\subsection{RL in Affective Computing} \label{sec:background-rl-affect}

In traditional affective computing, computational models of affect take the context of the user's interaction as input
---physiological signals \cite{martinez2013learning}, facial expressions \cite{ruiz2018multiinstance,walecki2017deep}, speech \cite{trigeorgis2016adieu}, or information on the stimulus itself \cite{makantasis2021pixels}--- and output a predicted corresponding emotional state (i.e. the ground truth of emotion). Typically, such models are trained in a supervised learning fashion on datasets consisting of user state-affect pairs \cite{calvo2010affect, Koelstra2012DEAPAD} where the user state includes the context and any considered affect manifestations. 
More recent literature incorporates learning models of affect using privileged information, which is information that is not available during testing, and has proved to significantly improve performance \cite{makantasis2021affranknet+}. These models of affect are then paired with action selection mechanisms to enable affect-driven behaviors, such as socially believable agents \cite{reilly1996believable}. In this paper we look beyond the traditional supervised learning paradigm and, similarly to \cite{barthet2021go}, we view affect modeling as a RL process.

Literature on RL and agent emotion mostly focuses on the effect of artificial/simulated emotion signals on training \cite{moerland2018emotion}, whereas the use of human-annotated emotion during training is relatively unexplored. However, combining an agent's simulated affect with its action-selection strategy can help an agent find its goal faster and prevent premature convergence \cite{broekens2007affect}, as well as teach agents to perform simple robot control tasks \cite{hasson2011emotions}. Recent work in emotion recognition has used RL for real-time training of multimodal emotion recognition models using videos, achieving state-of-the-art performance \cite{Zhang2022ERLDK}. The use of intrinsically motivated RL agents \cite{singh2005intrinsically,singh2010intrinsically} is also relevant to our work. Whilst intrinsic motivation in RL \cite{jaques2019social, hussenot2020show} abandons human training signals entirely, reward functions are inferred from behavior demonstrations through the inverse RL paradigm. Barthet \textit{et al.} \cite{barthet2021go} introduced \textit{Go-Blend} which extends Go-Explore in an effort to behave optimally whilst imitating human affect through the use of an arousal model based on human demonstrations to generate high quality trajectories, generating agents that imitate.

This work expands Go-Blend \cite{barthet2021go} significantly by exploring the performance of reward function variants for affect, and investigating the impact of arousal as a driver for exploration in Go-Explore. In contrast to the simple arcade game used in \cite{barthet2021go}, in this paper we validate Go-Blend using a challenging, fast-paced, and continuous-controlled racing game. 

\section{The Go-Blend Algorithm}\label{sec:goblend} 

\begin{figure}[!tb]
\centering
\includegraphics[width=.7\columnwidth]{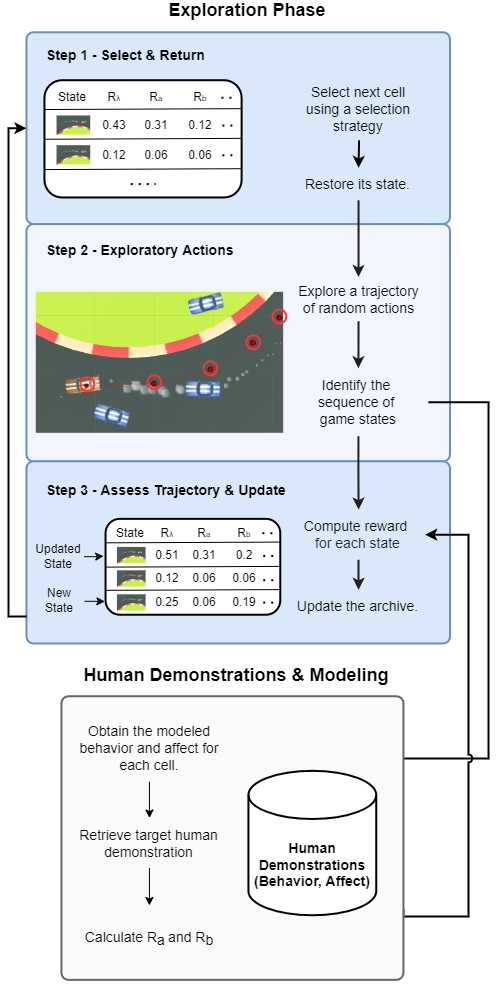}
\caption{A high-level overview of Go-Blend for modeling human demonstrations via RL. Go-Blend is capable of generating agents with predetermined behavioral and affective patterns as those are provided by reward functions}
\label{fig:algorithm}
\end{figure}

This section outlines the core aspects of Go-Blend \cite{barthet2021go}, the algorithm we build upon in this study. We first cover the high-level details of the Go-Blend algorithm (section~\ref{sec:algorithm}) and then we describe the different reward signals used in this study (section \ref{sec:rewards}). The section also outlines the ways we use affect traces to drive cell selection (section \ref{sec:selection}). Fig.~\ref{fig:algorithm} illustrates the core aspects of the Go-Blend framework.

\subsection{Algorithm} \label{sec:algorithm}

Go-Blend~\cite{barthet2021go} is an implementation of the Go-Explore RL algorithm~\cite{Go-Explore} which is able to generate agents that exhibit various behavioral and affective patterns as specified by its rewards functions. Due to the deterministic nature of the test-bed game we examine in this paper (see section \ref{sec:case_study}), we focus primarily on the exploration phase of the algorithm and, hence, we do not include the robustification phase in this implementation. This optional step creates an RL agent that is capable of performing at the level achieved during exploration in a stochastic environment.

During exploration, the system builds and maintains an archive of cells in an iterative fashion, with each cell representing a unique world state (e.g. a game state) that has been observed so far. Alongside their state, cells also store the trajectory of actions, previous states, and affect demonstrations (such as traces) required to reproduce their state. Each cell is assigned a behavior reward (e.g. maximize in-game score) denoted by $R_b$, and an affective reward (e.g. model annotated arousal) denoted by $R_a$. 

Exploration takes place by repeatedly selecting a cell from the archive using a cell selection strategy, loading its game state, and taking a number of exploratory actions to identify new cells. These exploratory actions are taken according to an action selection mechanism (e.g. random, domain knowledge, or policy-based). After each action, a cell is constructed according to the current game state, and is assigned rewards for both $R_b$ and $R_a$. New cells are always added to the archive if they have not been encountered so far. If the cell's state already exists in the archive, it is updated only if it satisfies one of two replacement criteria: a) any cell encountered with a better reward than its existing counterpart in the archive is updated, b) any cell with the same reward as the existing cell is updated if it has a more efficient (shorter) trajectory. New cells are selected for exploration from the archive according to a cell selection strategy in an iterative fashion. This process of selecting cells, returning to their state and exploring new actions is repeated for a fixed number of iterations or until a desired stopping criterion is reached (e.g. reaching the optimal score or exploring all possible states). The result of this exploration phase is a number of high-performing cells for the given deterministic environment.

\subsection{Reward Functions} \label{sec:rewards}

In this section, we describe in detail the different reward functions implemented and explored. Unlike the original Go-Blend case study \cite{barthet2021go}, we do not test a function which rewards behavior similarity (with or without affect). Instead, we focus on the impact of different reward functions on cell replacement and resulting agent behavior. The reward functions are designed under certain assumptions about the human demonstrations and model. First, we assume time-continuous signals for the phenomenon (e.g. behavior, manifestations of affect) being modelled. As a result, the Go-Blend trajectory produces a time-continuous signal, which should match a human-provided time-continuous target signal. We also assume that multiple human-provided signals are available to calculate deviations between them. 

Equation~\eqref{eq:similarity} defines our method for measuring the similarity between a data point in our Go-Blend trajectories and a data point in the target trace. This equation assumes the data points provided are normalized within $[0, 1]$.
\begin{equation}
S_a(i) =  \left(1 - |h_{a}(i)-t_{a}(i)|\right)^2
\label{eq:similarity}
\end{equation}
\noindent where $i$ is the time window being evaluated; $h_{a}(i)$ is the trajectory's affect value for time window $i$, and $t_{a}(i)$ is the target value respectively for time window $i$. The similarity is squared to penalize data points with low similarity more harshly, creating a more pronounced difference between low and high quality cells. This equation forms the basis for most reward functions in this section.

\subsubsection{Regression Reward ($R_{a}$)}

The first baseline function we explore for modeling affect attempts to match a target signal for affect via regression. Equation~\eqref{eq:predict_arousal} calculates the average similarity between the data points in the Go-Blend trajectory and the data points in the target affect trace. 
\begin{equation}
R_{a} = \frac{1}{n} \sum^{n}_{i=0} S_a(i)
\label{eq:predict_arousal}
\end{equation}
\noindent where $i$ is the time window being evaluated; $n$ is the number of time windows observed so far in this trajectory, and $S_a(i)$ is the similarity function defined in Eq.~\eqref{eq:similarity}. Since $R_a$ calculates the average difference across all the observations made so far, it encourages arousal trajectories with high imitation accuracy across the entire duration of the game.

\subsubsection{Uncertainty Reward ($R_{au}$)}

We expand upon $R_a$ by factoring in the uncertainty of the affect traces; uncertainty is expressed through multiple annotations of the same time window ($i$). The $R_{au}$ reward, in Eq.~\eqref{eq:arousal/std}, uses the standard deviation, i.e. $\sigma({S_a(i)})$, of the provided affect traces at time window $i$ as its uncertainty proxy thereby assisting the algorithm to prioritize states with high degrees of affect confidence.
\begin{equation}
R_{au} = \frac{1}{n}\sum^{n}_{i=0} {\frac{S_a(i)}{1 + \sigma({S_a(i))}}}
\label{eq:arousal/std}
\end{equation}
\subsubsection{Confidence Reward ($R_{ac}$)}

Our final reward function for modeling human affect signals views affect modeling from a different perspective. Rather than using similarity (Eq.~\ref{eq:similarity}) to a target affect trace as a reward signal, the $R_{ac}$ reward function is the average $r_{ac}$ of the action sequence where $r_ac$ returns a positive reward (+1) if the predicted affect lies within the 95\% confidence bounds of the target affect trace, and a negative reward ($-1$) otherwise (see Eq.~\ref{eq:arousal/confidence}). $R_{ac}$ penalizes trajectories which stray outside the confidence bounds of the target affect trace and pressures exploration to model the trace accurately.
%
\begin{equation}
r_{ac}(i)=\begin{cases}
            1 &  \text{if $|h_{a}(i) - t_{a}(i)| < c_{a}(i)$}  \\
            -1 & \text{otherwise}
        \end{cases}
\label{eq:arousal/confidence}
\end{equation}
\noindent where $c_{a}(i)$ is the 95\% confidence interval value for the target affect value at the same time window.

It is important to note that the aforementioned reward functions can model any form of manifestation of affect (e.g. arousal, valence, dominance) given appropriate traces. For more complex affect models, $h_{a}(i)$ and $t_{a}(i)$ can be seen as a vector for all the behavioral and affective dimensions that should be modeled. Section \ref{sec:solid_reward} provides an example of how these reward functions can be applied to our test-bed game.

\subsection{Cell Selection Strategy} \label{sec:selection}

A core focus of this paper is on the investigation of emotion as a facilitator of decision-making inspired by Damasio's \textit{somatic marker hypothesis} \cite{damasio1996somatic}. In practical terms, we test the impact of using affect reward ($R_a$) for cell selection on the exploration capacity of Go-Blend as opposed to a basic uniform selection method. Inspired by the weighted likelihood method for exploration \cite{Go-Explore}, we explore three different variants in this paper:

\subsubsection{Uniform Selection}\label{sec:selection_uniform}

The default cell selection strategy for both Go-Explore and Go-Blend is to select randomly, with equal chance, among the cells that have already been added to the archive. We use this uniform selection as our baseline.

\subsubsection{Roulette-Wheel Selection}\label{sec:selection_roulette} 
This cell strategy simply uses the reward for affect ($R_a$ of Eq.~\ref{eq:predict_arousal} in this paper) as the selection likelihood for each cell ($i$), which we call $W_a$ (shown in Eq.~\ref{eq:roullete}). Selecting by $W_a$ can be thought of as a straightforward roulette-wheel~\cite{lipowski2012roulette} selection where we bias cell selection to exploit high-performing cells in terms of affect modeling.
\begin{equation}\label{eq:roullete}
    W_{a}^i = \frac{R_a^i}{\sum_{j=1}^{n} R_a^j}
\end{equation}
\noindent where $W_{a}^i$ is the probability of selecting cell $i$ in the archive, $n$ the number of visited cells in the archive, $R_{a}^{i}$ the reward of the action sequence resulting in the game state of cell $i$.

\subsubsection{UCB Selection}\label{sec:selection_ucb} 
This cell selection strategy combines $W_a$ with a selection weight inspired by the Upper Confidence Bound (UCB) selection policy implemented in Go-Explore \cite{Go-Explore} (see Eq.~\ref{eq:selection}). This selection weight, namely $W_e$, balances exploitation of high reward cells (i.e. which accurately model human affect) with exploration by reducing the probability of a cell being selected if it has been visited many times.
\begin{equation}
W_{e}^i =  \frac{W_a^i}{\sqrt{C_{seen}^i+1}}
\label{eq:selection}
\end{equation}
\noindent where $C_{seen}^i$ is the number of times cell $i$ has been selected for exploration.

\section{Test-Bed Racing Game: Solid Rally} \label{sec:case_study}

To test the different selection and reward functions for modeling affect traces of human players, we use the ``Solid Rally'' driving game (hereafter \emph{Solid}) which forms part of the AGAIN dataset \cite{AGAIN}. In this section, we describe the properties of the game and the accompanied affect dataset.

\subsection{Game Description} \label{sec:case_study_game}

\begin{figure}[!tb]
\centering
\includegraphics[width=0.89\columnwidth]{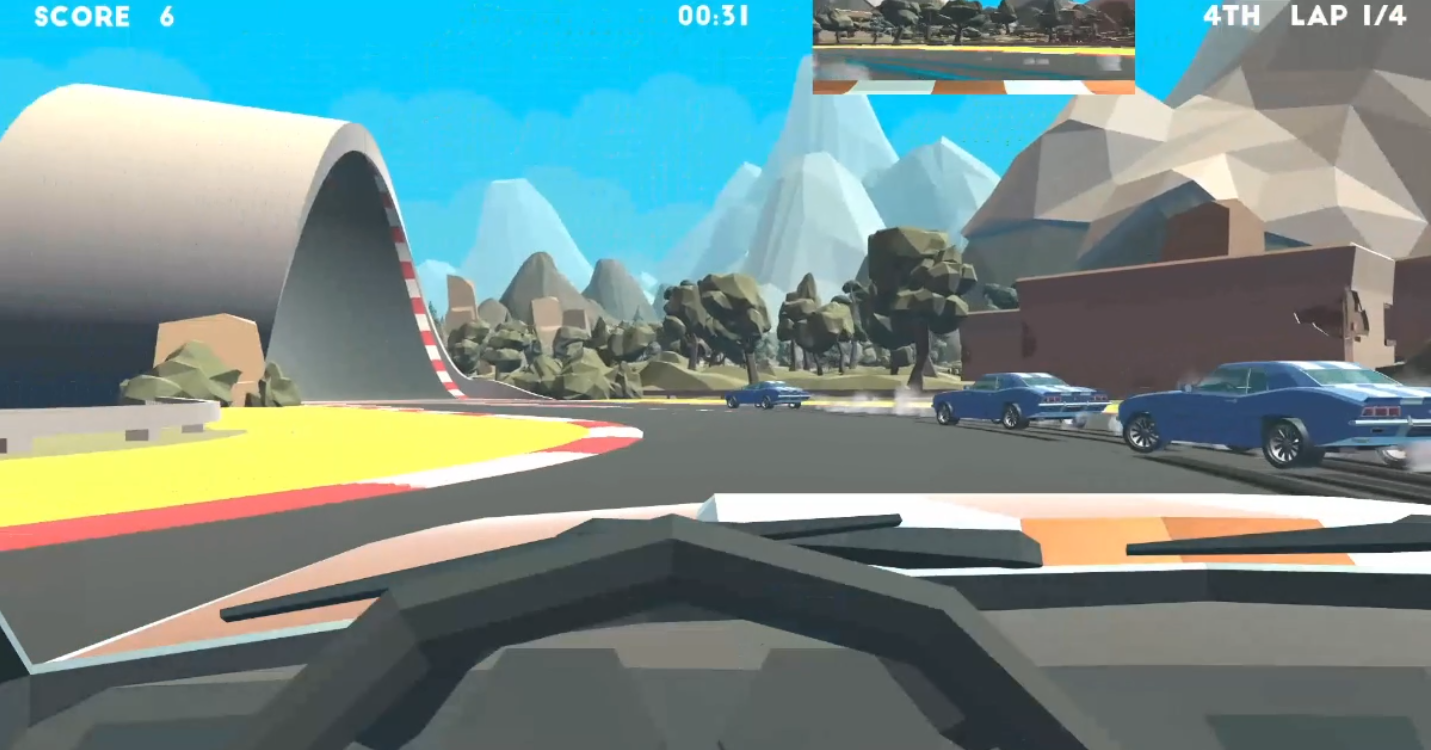}
\caption{First-person view in the Solid Rally racing game.}
\label{fig:solid}
\end{figure}

In Solid, the player controls a rally car from a first-person perspective (see Fig.~\ref{fig:solid}) and attempts to win a race against three opponent cars. The race ends when the player has completed three laps or after two minutes. The player accumulates points by successfully driving around the circuit and passing through checkpoints (eight per lap) for a maximum score of 16 points awarded for completing the full two laps within 90 seconds. 
The player controls the car's gas pedal and steering wheel through the arrow keys. There are three possible inputs for steering (-1, 0, 1) and gas (-1, 0, 1), where 0 is the neutral state when no key is pressed. The track contains ``off-track'' grass segments which slow the cars down slightly if driven over, providing a viable strategy to cut paths through the curves in the circuit. Opponent cars are driven by AI via a simple deterministic controller.

\subsection{AGAIN Dataset}\label{sec:again_set}

Solid Rally is accompanied by a dataset of 108 human play sessions (after removing outliers), each containing player annotated arousal traces \cite{AGAIN}. Each play session consists of 250ms time windows containing the average data for 32 in-game features specific to Solid. These features cover basic spatial properties such as rotation, speed, and collision status for both the player and opponent cars, as well as their score. Arousal traces were collected in a continuous, unbounded fashion using RankTrace \cite{RankTrace} and the PAGAN \cite{PAGAN} online annotation framework. These arousal traces were then normalized on a per-session basis to account for discrepancies in the value ranges between players. 

Note that due to the large size of the dataset, and a significant gap in performance between players, we used hierarchical clustering on players' aggregated session data to identify the best performing group. Details on the clustering process and experiments with Go-Blend on different player groups are provided in \cite{barthet2022solidpersonas}. The result of this process is a dataset of 27 ``expert'' players with similar behavioral characteristics. All of these expert players complete the 2 laps within at least $88.25$ seconds (353 time windows) and all reach the maximum score (of 16) within that timeframe. The mean arousal trace of these expert players is used as our target ($t_a(i)$ in Eq.~\ref{eq:similarity}) for modeling affect.

\section{Go-Blend for Solid Rally} \label{sec:case_study_method}

In this section, we focus on the implementation details of Go-Blend when employed in the Solid Rally test game. In particular, we explore how game states are stored in the archive (Section~\ref{sec:solid_state}) and how the different rewards are calculated for the game (Section~\ref{sec:solid_reward}).

\subsection{State Representation}\label{sec:solid_state}

A critical design choice for a Go-Explore algorithm such as Go-Blend is the method for mapping game states to cells to be stored in the archive. The game-specific features, provided by AGAIN \cite{AGAIN}, lack sufficient granularity to adequately distinguish between meaningfully different game states, therefore a new cell state representation was needed. Our chosen state representation is determined by 6 categorical variables. The first two are defined by the player car's \textit{speed}---comprising two states (slow, fast)---and \textit{rotation}---comprising six 30-degree thresholds. To identify the car's location, the circuit is split into 19 high-level segments which are further subdivided into sub-segments according to their shape (e.g. left side, right side, right off-road). The last two variables consider the player's current lap number, and the player's \textit{proximity} to opponent cars---i.e. an opponent car is on the same sub-segment as the player or not. This game state representation has a maximum archive size of $4,800$ cells, or $2,400$ cells per lap.

\subsection{Reward Calculation}\label{sec:solid_reward}

We use the annotated arousal traces provided with the human demonstrations as our ground truth. Note that we directly use AGAIN's arousal traces to build our reward functions rather than relying on a trained surrogate model of arousal to predict outcomes indirectly. Our approach for calculating the arousal value for a cell, i.e. $h_{a}(i)$ in Equation~\eqref{eq:similarity}, is similar to that used by Barthet \textit{et al.}~\cite{barthet2021go}. 

After each exploratory action taken, the algorithm queries the dataset of human demonstrations for the arousal values of the $k$-nearest neighbors. The arousal at the current time window $i$ ($h_{a}(i)$), is taken as the mean of these arousal values, calculated using distance-weighted $k$-NN \cite{dudani1976distance}. We find the $k$ nearest neighbors ($k=5$ throughout this paper) through the Euclidean distance from the current feature vector (i.e. the vector of 32 features) to every entry in the dataset of demonstrations. To find the weighted average, each arousal value is weighted by their corresponding distance to give less weight to entries further away from the current state. These weighted values are added together and divided by the sum of the neighbor distances to obtain a value normalized within $[0,1]$.
Using distance weighted $k$-NN helps reduce the noise caused by outliers in the human playtraces and biases arousal similarity to the arousal values coming from the most similar game-states at each time window. Note that the target arousal for each time window ($t_a(i)$) is calculated from the mean arousal trace value for all expert players.
The $k$ nearest neighbors are also used for the uncertainty reward ($R_{au}$). We calculate the standard deviation between the $k$-NN arousal values and use it as the uncertainty penalty ($\sigma({S_a(i)})$ in Equation~\ref{eq:arousal/std}) at time window $i$.

\begin{table*}[!tb]
\caption{
Results for the arousal reward functions used for cell replacement in Solid Rally, and averaged across 3 runs, including the 95\% confidence interval. Boldface values point to the highest value (or lowest, for uncertainty) across agents.
} 

\begin{center} 
\begin{tabular}{|c||c|c||c|c|c|c||c|}
\hline
\textbf{Arousal} & \multicolumn{2}{c||} {\textbf{In-Game Statistics}} & \multicolumn{4}{c||} {\textbf{Arousal Modeling Performance Measures}} & \multicolumn{1}{c|} {\textbf{Go-Blend Performance}}\\
\cline{2-8} 
\textbf{Reward} & Final Score & Behavior CCC & Mean Arousal & Arousal CCC & Arousal Deviation & Confidence & Archive Size (\%)  \\
\hline
\hline
Random Agent & 0.00$\pm$0.00 & 0.00$\pm$0.00 & 0.33$\pm$0.08 & 0.12$\pm$0.09 & 0.17$\pm$0.01 & $-0.54$$\pm$0.24 & N/A \\
\hline
Max Score & \textbf{12.33$\pm$0.53} & \textbf{0.83$\pm$0.02} & 0.39$\pm$0.02 & 0.11$\pm$0.04 & 0.17$\pm$0.01 & -0.47$\pm$0.10 & 71.74$\pm$2.23 \\
\hline
Max Arousal & 3.00$\pm$2.44 & 0.18$\pm$0.15 & \textbf{0.62$\pm$0.00} & 0.10$\pm$0.11 & 0.17$\pm$0.01 & -0.59$\pm$0.02 & 46.58$\pm$0.05\\
\hline
\hline
$R_a$ & 11.00$\pm$2.44 & 0.81$\pm$0.12 & 0.50$\pm$0.01 &  \textbf{0.75$\pm$0.01} & 0.14$\pm$0.01 & 0.33$\pm$0.04  & \textbf{82.34$\pm$10.36} \\
\hline
$R_{au}$ & 9.33$\pm$2.13 & 0.69$\pm$0.21 & 0.50$\pm$0.01 & 0.72$\pm$0.07 &\textbf{0.12$\pm$0.02} & 0.29$\pm$0.00 & 75.01$\pm$15.46 \\
\hline
$R_{ac}$ & 8.00$\pm$0.00 & 0.52$\pm$0.05 & 0.48$\pm$0.00 & 0.61$\pm$0.09 & 0.15$\pm$0.01 & \textbf{0.54$\pm$0.08} & 59.70$\pm$3.34 \\
\hline
\end{tabular}
\label{tab:reward_table}
\end{center}
\vspace{-10pt}
\end{table*}

\section{Experiments} \label{sec:results}

To investigate the impact of affect-based reward functions in Go-Blend and the impact of arousal as a selection mechanism for effective RL search, we perform two sets of experiments: we first employ the different reward functions defined in Section \ref{sec:rewards} for \emph{cell replacement} and attempt to match players' arousal states (see Section \ref{sec:cellReplacement}), and then explore the three different arousal-based selection strategies for \emph{cell selection} towards better agent performance (see Section \ref{sec:cellSelection}).

Each experiment reported in this section is conducted with 3 independent runs for $500,000$ iterations of exploration, saving the cell with the best reward value. During an iteration of exploration, 25 exploratory actions are taken before randomly selecting a new cell to load and explore from. Exploratory actions are selected every 250ms using a weighted random selection based on the average action frequency logged by our player database. 

We compare all our generated agents against three baseline agents: a \textit{random agent} that chooses a gas/steering action via weighted random selection (see above), a \textit{max score} agent using Go-Blend to maximize its score (in-game behavior) within the time limit, and a \textit{max arousal} agent that uses Go-Blend to maximize the arousal value during the whole session.

We use the following performance measures to assess how our various Go-Blend agents play Solid Rally for 90 seconds. To assess in-game performance, we measure the agent's \textit{final score}. To assess how accurately the experiments are able to model the given traces for behavior and affect, we include the \textit{Concordance Correlation Coefficient} (CCC) between experiments' trajectories and target (player) trace for behavior and affect \cite{CCC_Metric}. We measure \textit{arousal deviation} (standard deviation of similarity between $k$-NN player traces, averaged throughout the playthrough) and \textit{confidence} ($R_{ac}$ calculated at the end of the playthrough) which are used for our rewards in section \ref{sec:rewards}. We also measure the \textit{mean arousal} value over the whole session. Finally, we include the \textit{archive size} as a percentage of the maximum possible cells that can be encountered, in order to assess the exploration capacity of the algorithm. 

\begin{figure}
\centering
\includegraphics[width=\columnwidth]{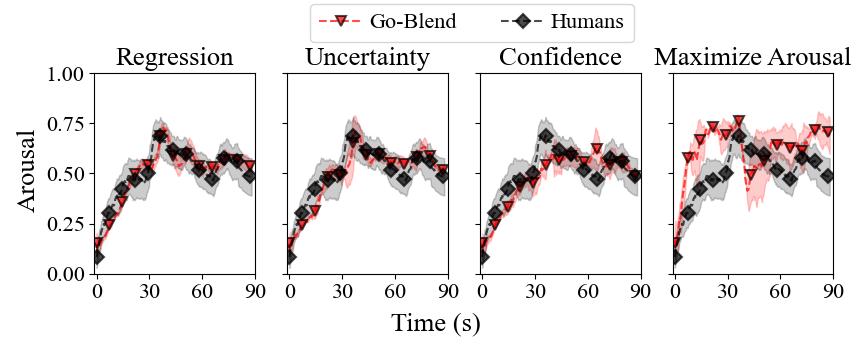}
\caption{Moving averages for arousal traces of the best trajectories generated by Go-Blend when using arousal-driven reward functions (see section \ref{sec:rewards}). Results are averaged from 3 runs and include the 95\% confidence interval. }
\vspace{-7pt}
\label{fig:arousal_traces}
\end{figure}

\subsection{Cell Replacement} \label{sec:cellReplacement}

Table \ref{tab:reward_table} shows a set of performance measures for three Go-Blend experiments and three baseline experiments. The random agent clearly struggles on all fronts, while the \textit{max score} agent reaches a decent final score within the time limit (considering that all players' score is 16 in the dataset), likely due to the sparseness of the rewards. As expected, \emph{max arousal} causes the agent to increase their perceived arousal and does not match the players' arousal (low arousal CCC) and performs poorly. In terms of correlation to the mean human arousal trace, the regression ($R_a$), uncertainty ($R_{au})$ and confidence ($R_{ac})$ experiments produce much better results than the baselines. Moreover, we observe that the explicit targets (uncertainty, confidence) are reached more optimally by the reward structures that target them ($R_{au},R_{ac}$ respectively). Note that we aim to minimize arousal deviation for a more consistent playthrough. Looking at the arousal traces of Fig.~\ref{fig:arousal_traces}, we can clearly see the capabilities of Go-Blend for modelling affect. All three experiments which model the mean human arousal trace exhibit arousal traces which lie within, or just outside, the 95\% confidence interval of the target trace. 

While arousal metrics are optimally reached by the reward structures that explicitly target it (arousal CCC by $R_a$, minimal deviation by $R_{au}$, confidence by $R_{ac}$), these rewards also result in different in-game behaviors. Of the three reward variants, $R_a$ performs better, reaching performance almost at the level of the agent that targets in-game score. Moreover, it better aligns with in-game behavior of actual players (behavior CCC). Importantly, using $R_a$ for cell replacement results in more efficient exploration of the Go-Blend archive size, and this likely causes the algorithm to explore from more states and thus perform well overall. The $R_{au}$ performs adequately in terms of score, and also fills the state-space archive better than the max score agent. Surprisingly, due to the efficient exploration of these experiments, both archives for $R_a$ and $R_{au}$ contain at least one cell with a higher final score than the max score agent ($13.67$). Meanwhile, $R_{ac}$ does not result in well-performing agents (low final score and behavior CCC) and has a similar drop in archive size. We hypothesize that the binary reward mechanism used by $R_{ac}$ provides sparser arousal rewards than the other two variants, and thus stifles exploration which, in turn, stifles performance.

\begin{figure}[t]
\begin{minipage}{\columnwidth}\centering
\includegraphics[width=0.9\textwidth]{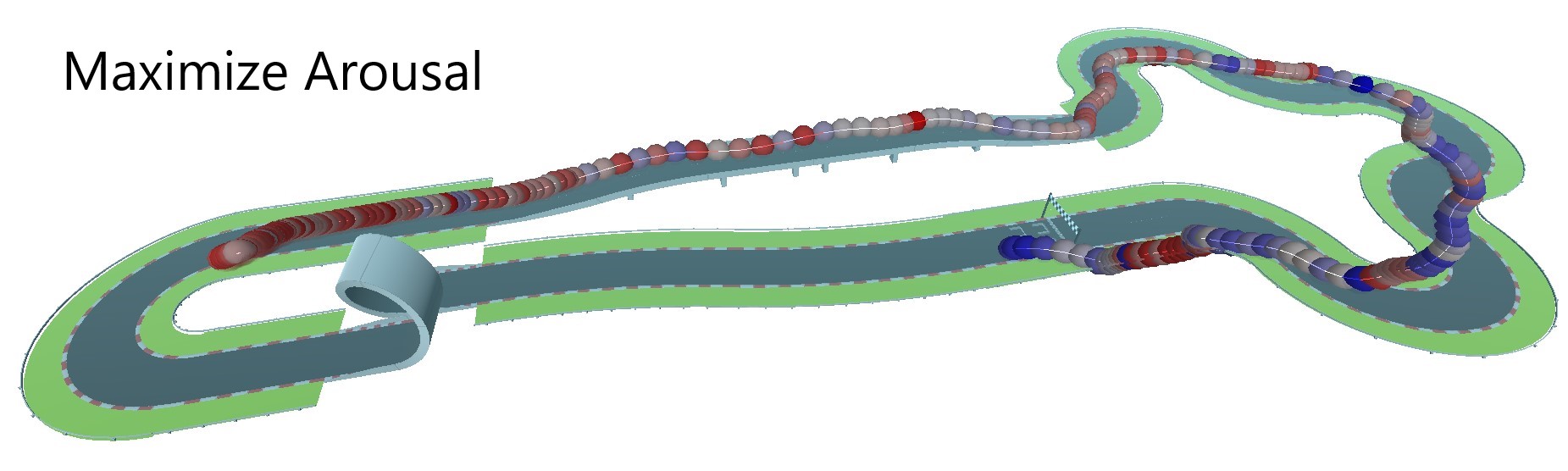}
\end{minipage}
\begin{minipage}{\columnwidth}\centering
\includegraphics[width=0.9\textwidth]{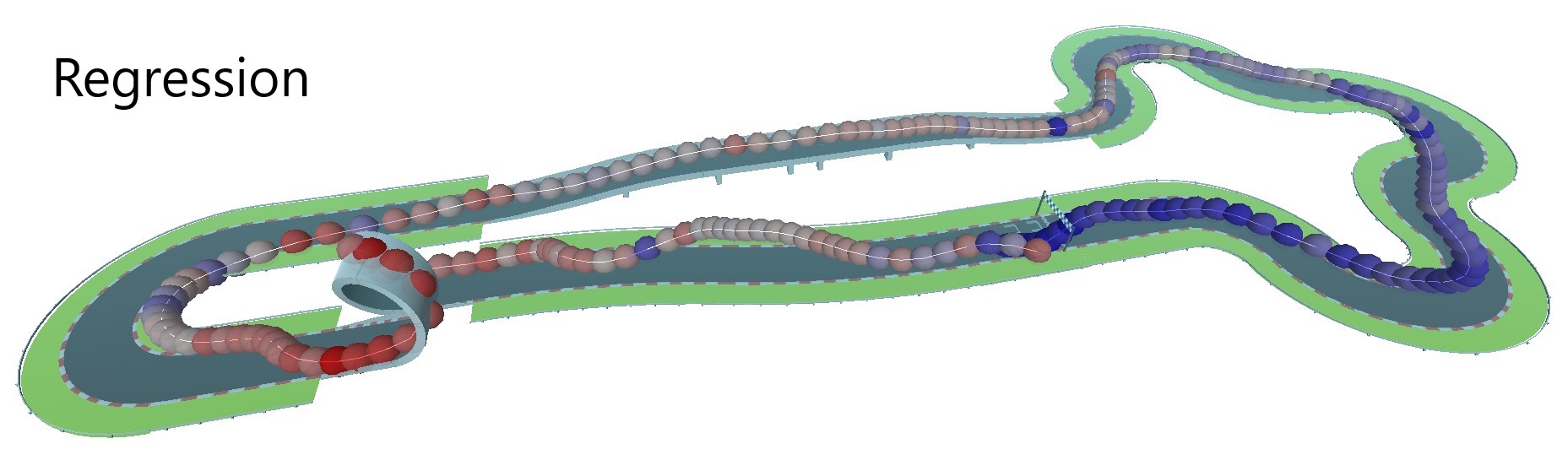}
\end{minipage}
\begin{minipage}{\columnwidth}\centering
\includegraphics[width=0.9\textwidth]{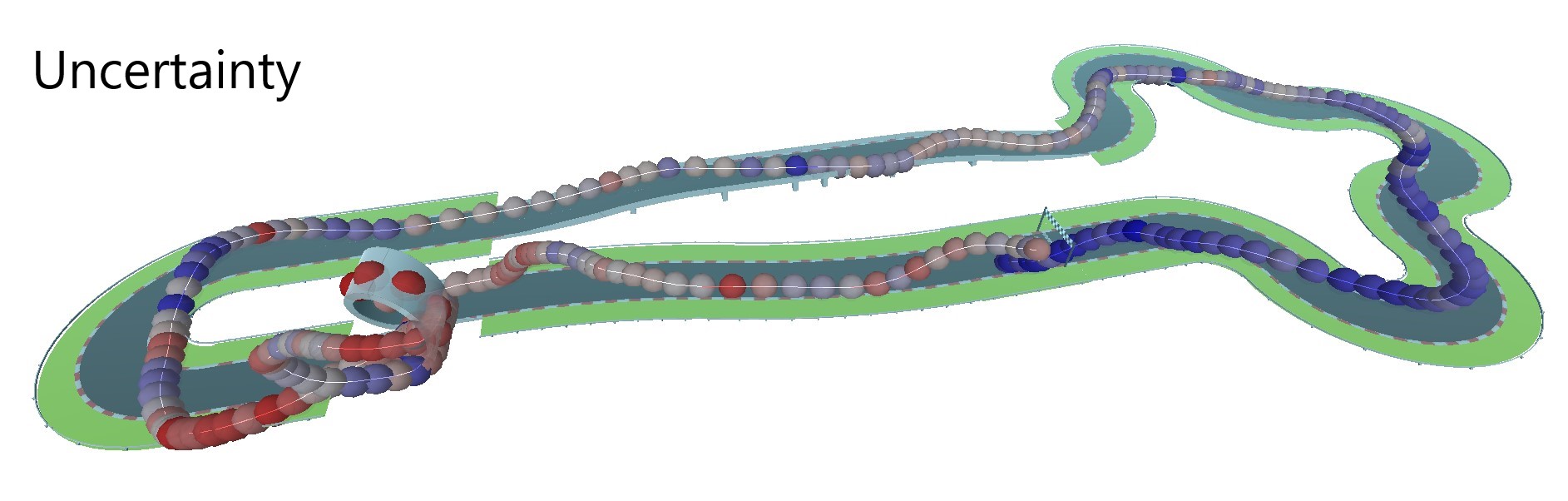}
\end{minipage}
\begin{minipage}{\columnwidth}\centering
\includegraphics[width=0.9\textwidth]{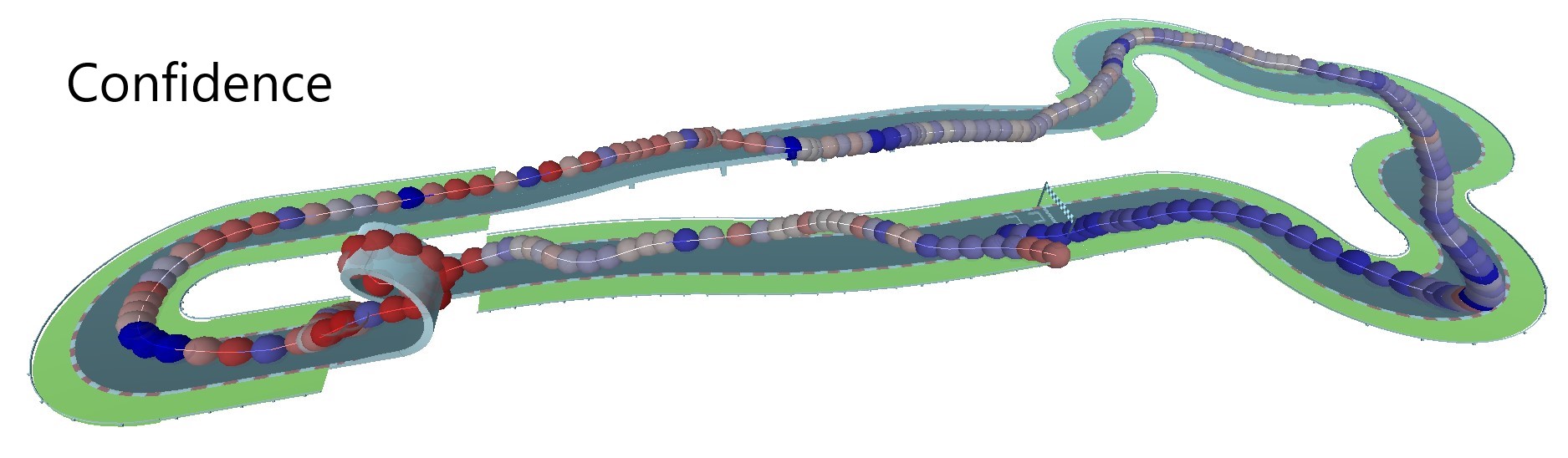}
\end{minipage}
\caption{The best playthroughs among 3 runs of the trained agents for different arousal rewards. Spheres indicate the player's position during the first lap, with red and blue spheres indicating high and low predicted arousal respectively.}
\vspace{-7pt}
\label{fig:solid_traces}
\end{figure}

We show indicative playthroughs of agents trained via arousal-based rewards in Fig.~\ref{fig:solid_traces}, for the first lap of a Solid race. We observe that when maximizing arousal, the agent does not even finish the first lap. Interestingly, we also observe erratic behavior, with abrupt changes in direction even when the car's trajectory seems good. This results in more time spent offroad (60\% on average); we expect that users annotate portions of their playthrough where they make mistakes as highly arousing (i.e. frustrating), and therefore the agent attempts to recreate such mistakes by going offroad and cutting corners at all times. When trained based on $R_a$, the agent drives on the road and takes good turns throughout the level. Moreover, while the max arousal agent finds arousing states even in straight parts of the level (see bridge portion), the agent trained on $R_a$ only finds arousing moments when performing the gravity-defying loop---which matches our own intuitions as players. Agents trained via $R_{au}$ and $R_{ac}$ perform somewhere in-between the above two extremes, with $R_{au}$ making more mistakes when approaching the loop and $R_{ac}$ making mistakes at the straight bridge section. We also observe that the $R_{ac}$ agent overall has fewer sections with high arousal (see mean arousal of Table \ref{tab:reward_table}) and its low game score seems to be attributed to slow driving (average speed of 25 units/s compared to 33 units/s for $R_a$) rather than to mistakes (16\% time offroad on average versus 32\% for $R_a$). 

\begin{table}[!tb]
\caption{
Results for the arousal driven cell selection strategies averaged across 3 runs, with 95\% confidence intervals. The reward function is maximizing game score rewards. Boldface values point to the highest value across agents.
} 
\begin{center} 
\begin{tabular}{|c|c|c|c|}
\hline
Selection &  Final Score &  Behavior CCC & Archive Size (\%) \\
\hline
\hline
{Uniform} & 12.33$\pm$0.53 & 0.83$\pm$0.02 & 71.74$\pm$2.23 \\
\hline
$W_a$ & \textbf{13.33$\pm$0.53} & \textbf{0.86$\pm$0.02} & \textbf{73.03$\pm$3.07} \\
\hline
$W_e$ & 12.67$\pm$0.53 & 0.81$\pm$0.03 & 72.35$\pm$2.29 \\
\hline
\end{tabular}
\vspace{-5pt}
\label{tab:selection_table}
\end{center}
\end{table}

\subsection{Cell Selection} \label{sec:cellSelection}
We explore how cell selection biased by conformance to arousal (see section \ref{sec:selection}) affects a well-performing agent in terms of game behavior. We only focus on the max score agent in these experiments, and report how its playtrace fares after training biased by different cell selection strategies in Table \ref{tab:selection_table}. We observe that while values of the behavioral metrics are fairly similar, biasing with either $W_a$ or $W_e$ leads to better final scores and exploration of the state space (archive size), especially so for $W_a$. Trying to explain such behavior through gameplay metrics, we observe that the agents trained via $W_a$ and $W_e$ retain the same average speed as with uniform selection, while staying more on the road (34\% of time offroad for uniform, versus 28\% for $W_a$ and 26\% for $W_e$). Our findings suggest that the offroad and midair (loop) sections of the track contain states with arousal values which are easier to predict. A Pearson correlation analysis showed that offroad and midair states are consistently annotated with higher arousal values than regular states with correlation values of $0.38$ and $0.39$, respectively. We assume that the bias to perform more (random) playthroughs in situations where conformance with players' arousal is high leads to more robust behaviors in those segments. Therefore, performing more simulations in those sections lets the agent become more competent at taking turns in a better fashion without sacrificing speed (unlike e.g. the $R_{ac}$ agent in Fig.~\ref{fig:solid_traces}).

\section{Discussion} \label{sec:discussion}

As discussed in the introduction, we assume that leveraging affect states to bias the training process of an RL agent will act as a facilitator for decision-making. Testing our hypothesis on a racing game, we observe that---unsurprisingly---agents that receive only behavioral rewards (playing to win) tend to take better decisions towards achieving higher in-game scores. However, biasing where the RL agent explores from according to a similarity to expert players' arousal signals can lead to better behaviors for the---traditional---RL agent that plays to win. On the other hand, imitating expert players' arousal resulted in good in-game behavior, but not as high-performing as behavior-based rewards. Using this arousal consensus also led to better exploration potential for the algorithm itself. 

It should be noted that this paper did not use all players of Solid Rally that are contained in the AGAIN dataset \cite{AGAIN}, but instead focused only on high-performing players. Interestingly, imitating these players' arousal traces biased the agents towards correct in-game strategies. Experiments with low performing players \cite{barthet2022solidpersonas} did not result in competent agents when imitating arousal alone, which means that conclusions regarding the power of arousal as a reward or strategy in RL tasks pre-supposes that high-quality arousal and performance examples exist. On the other hand, the fact that only a small portion of the population is considered may explain why reward based on consensus between arousal traces ($R_{au}$, $R_{ac}$) did not perform as well. Using a larger player base with less consistent behaviors and corresponding arousal signals may give more meaning to notions such as `uncertainty' compared to the few and fairly uniform behavior/affect traces of expert players. Other hyperparameters that may affect the impact of the rewards and Go-Blend more broadly are the number of nearest players considered ($k$ in the $k$-NN algorithm used), as well as how game-states are matched. In this paper, we follow the literature \cite{barthet2021go} and use high-level ad-hoc designed features that describe the game according to the nuances of its genre, but other options such as similarity of the annotation stimulus (i.e. the recorded game footage) or a game-state similarity weighted by the overall performance similarity between the two players (similar to how we derived the expert player cluster) could yield different strategies.

This paper extends the original implementation of Go-Blend which was validated in Endless \cite{barthet2021go}, a simple game with a small action space. Through Solid Rally, we test a more complex game that can trigger more emotionally rich reactions, and a broad variety of behaviors---which necessitated our aforementioned filtering of expert behaviors. However, testing how Go-Blend performs in different games with more complex mechanics (e.g. first-person shooters) and in-game behavioral rewards beyond score (e.g. role-playing games) is necessary to identify its strengths and limitations. 
It remains, for instance, an open question how such time-continuous traces and their imitation via RL can scale in longer games such as strategy games. Finally, an important limitation of experiments conducted within Endless \cite{barthet2021go} and Solid Rally is that the environment is deterministic and thus can be recreated faithfully whenever exploring from any cell in the Go-Blend archive. Stochastic game-state transitions would likely lead to game-states where no clear ``neighbor'' exists among human traces and therefore arousal imitation may become significantly more noisy.
Incorporating the robustification phase of Go-Explore algorithm \cite{Go-Explore} into Go-Blend is expected to lead to new insights on the impact of affect-based rewards, especially those based on uncertainty, as well as allow for the algorithm to scale to more complex games.

\section{Conclusions} \label{sec:conclusion}

In this paper we viewed affect modeling from an RL perspective and tested the hypothesis that RL processes can blend affect elicitation and manifestation with behavioral and affective responses efficiently. Inspired by the somatic marker hypothesis \cite{damasio1996somatic}, we also hypothesized that emotion can be the driver and facilitator of decision-making. 
We used behavioral and arousal demonstrations from a challenging racing game and tested our hypotheses by employing the Go-Blend RL algorithm and modeling affect through various reward functions. 
In addition, we tested how optimal game behaviors could be influenced by using the accuracy of the arousal model as a driver of search for the RL algorithm. Our findings indicate that when choosing states to explore based on their potential for modeling arousal can lead to better in game performance. Interestingly, modeling arousal based on the expert players' annotations led to high-performing agents despite the fact that ``winning'' the game was not explicitly targeted by the algorithm. Along with \cite{barthet2022solidpersonas}, this is the first application of the Go-Blend algorithm in a complex and affect-intense game, opening many directions for future work at the intersection of affect modeling, RL and affect-driven machine learning.

\bibliographystyle{IEEEtran}
\bibliography{main}

\end{document}